\pdfoutput=1
\documentclass[11pt]{article}
\usepackage{ACL2023}
\usepackage{times}
\usepackage{latexsym}
\usepackage{ulem} 
\usepackage{graphicx} 
\usepackage{multirow}
\usepackage[T1]{fontenc}
\usepackage[utf8]{inputenc}
\usepackage{microtype}
\usepackage{inconsolata}
\title{Towards a Common Understanding of Contributing Factors for Cross-Lingual Transfer in Multilingual Language Models: A Review}

\author{Fred Philippy\textsuperscript{1,2}\thanks{ \hspace{0.1cm} Research was conducted at Zortify.} \and Siwen Guo\textsuperscript{1} \and Shohreh Haddadan\textsuperscript{1} \\ \\ \textsuperscript{1}Zortify Labs, Zortify S.A. \\ 19, rue du Laboratoire L-1911 Luxembourg \\ \textsuperscript{2}SnT, University of Luxembourg \\ 29, Avenue J.F Kennedy L-1359 Luxembourg \\ \texttt{\{fred, siwen, shohreh\}@zortify.com}}

\begin{document}
\maketitle
\begin{abstract}
In recent years, pre-trained Multilingual Language Models (MLLMs) have shown a strong ability to transfer knowledge across different languages. However, given that the aspiration for such an ability has not been explicitly incorporated in the design of the majority of MLLMs, it is challenging to obtain a unique and straightforward explanation for its emergence. In this review paper, we survey literature that investigates different factors contributing to the capacity of MLLMs to perform zero-shot cross-lingual transfer and subsequently outline and discuss these factors in detail. To enhance the structure of this review and to facilitate consolidation with future studies, we identify five categories of such factors. In addition to providing a summary of empirical evidence from past studies, we identify consensuses among studies with consistent findings and resolve conflicts among contradictory ones. Our work contextualizes and unifies existing research streams which aim at explaining the cross-lingual potential of MLLMs. This review provides, first, an aligned reference point for future research and, second, guidance for a better-informed and more efficient way of leveraging the cross-lingual capacity of MLLMs.
\end{abstract}

\section{Introduction}
The objective of cross-lingual transfer is to leverage knowledge learned by a model in a source language and to transfer it to a target language. While such a process of transferring knowledge and concepts across languages seems natural for a polyglot, it is believed to be less straightforward for a language model. Nevertheless, multilingual language models (MLLMs), such as mBERT \citep{devlin_bert_2019}, XLM \citep{conneau_cross-lingual_2019} and XLM-R \citep{conneau_unsupervised_2020} demonstrate effective cross-lingual transfer capabilities. Such a transfer ability is moderately expected from XLM, given that parallel data is leveraged through a cross-lingual transfer learning objective during pre-training. However, it is less anticipated for mBERT and XLM-R, which are pre-trained on separate monolingual corpora without any explicit cross-lingual signal. Nevertheless, the latter show a surprisingly strong cross-lingual transfer capacity on a variety of downstream tasks \citep{hu_xtreme_2020}. While no apparent factors explaining the nature of this ability can be intuitively derived from the properties of MLLMs, there have been many attempts to understand this behavior. Past research has outlined and investigated various factors that may impact cross-lingual transfer performance in MLLMs, but there are still open questions due to conflicting findings across studies.
In our work, we inspect findings from past research investigating the inner workings of cross-lingual transfer in MLLMs. We not only outline overlapping contributions with consensual findings but also highlight and attempt to resolve conflicts between contradictory studies. Our work is structured according to five different types of factors whose impact on cross-lingual transfer capacity has been investigated in the past:

\begin{enumerate}
    \item Linguistic Similarity
    \item Lexical Overlap
    \item Model Architecture
    \item Pre-Training Settings
    \item Pre-Training Data.
\end{enumerate}

The examination of these factors provides insight into how and why MLLMs perform differently in different contexts. This understanding contributes to the overall explainability of MLLMs, which is essential for efficiently leveraging their cross-lingual transfer capacities and improving their performance in general. 

A list of all the papers surveyed in this study is provided in Appendix \ref{sec:appendix}.

\section{Background}
\subsection{Multilingual Language Models}
State-of-the-art MLLMS are predominantly based on the Transformer architecture \citep{vaswani_attention_2017}. These models aim to produce multilingual representations of text that can be used for various downstream tasks across different languages. However, MLLMs may adopt different learning objectives to achieve this goal. Some models exploit parallel data and incorporate a cross-lingual learning objective during pre-training, such as XLM \citep{conneau_cross-lingual_2019} and UniCoder \citep{huang_unicoder_2019}, while other models rely on separate monolingual corpora without any explicit cross-lingual supervision, such as mBERT \citep{devlin_bert_2019} and XLM-R \citep{conneau_unsupervised_2020}.

Despite their impressive performance, MLLMs also face several challenges and limitations, such as the imbalance in the pre-training data, the limited availability of evaluation datasets for different (low-resource) languages and the trade-off between model capacity and language coverage, known as the \textit{curse of multilinguality}, which affects their efficiency and effectiveness.. Therefore, more research is needed to understand, improve, and develop multilingual models that can achieve a balanced and robust performance across languages. Within this line of research, cross-lingual transfer has proven to be a valuable method to leverage resources from high-resource languages to improve downstream task performance for low-resource languages.

\subsection{(Zero-Shot) Cross-Lingual Transfer} \label{cl_transfer_def}
In the context of MLLMs, cross-lingual transfer refers to transferring certain knowledge from one language to another. From a practical standpoint, a traditional pipeline for zero-shot cross-lingual transfer typically includes two steps: \textbf{i)} A multilingual model is fine-tuned on a labeled dataset in the source language, and \textbf{ii)} The fine-tuned model is applied to a target language without any additional fine-tuning. In a few-shot setting, a small number of labeled samples in the target language are utilized for additional fine-tuning of the model.

During recent years, a number of studies have investigated cross-lingual transfer methods \citep{pikuliak_cross-lingual_2021}. In addition to the zero-shot transfer approach, there are some studies that apply machine translation to enable cross-lingual transfer \citep{conneau_xnli_2018, conneau_cross-lingual_2019, conneau_unsupervised_2020, hu_xtreme_2020}. In the translate-train approach, the labeled training set is translated from the source language into the target language for the purpose of fine-tuning. Correspondingly, the translate-test approach involves translating the test set from the target language into the source language during inference. In our review, we focus on the aforementioned traditional cross-lingual transfer process to avoid making the assumption that a translation system for the source language is available. Additionally, given that machine translation is highly context-dependent and is often unreliable when dealing with unconventional and ambiguous languages, it would add external factors to our effort of trying to understand the transfer behavior of MLLMs.

\section{Factors That Affect Cross-Lingual Transfer} \label{factors}

\subsection{Linguistic Similarity}
\begin{table*}[t]
    \small
    \centering
    
\renewcommand{\arraystretch}{1.5} 
\begin{tabular}{p{3cm}p{2cm}p{2.7cm}p{1cm}p{3cm}p{1.5cm}}
\hline
\multirow{2}{*}{\textbf{Paper}} & \multirow{2}{*}{\textbf{Task}} & \multirow{2}{*}{\textbf{Model}} & \textbf{Lang. type} & \multirow{2}{*}{\textbf{Features}} & \multirow{2}{*}{\textbf{Metric}} \\
\hline
\multirow{2}{*}{\citet{lin_choosing_2019}} & DP, EL, MT, POS & \multirow{2}{*}{/} & \multirow{2}{*}{NL} & GEN, GEO, INV, PHON, SYN & \multirow{2}{*}{lang2vec} \\
\hline
\citet{pires_how_2019} &  NER, POS & mBERT & NL & SYN & WALS \\
\hline
\citet{tran_zero-shot_2019} & DP & mBERT & NL & SYN & lang2vec \\
\hline
\citet{dufter_identifying_2020} & \multirow{2}{*}{SR, WA, WT} & \multirow{2}{*}{BERT (small)} & \multirow{2}{*}{SL} & \multirow{2}{*}{SYN} & \multirow{2}{*}{/} \\
\hline
\citet{k_cross-lingual_2020} & NER, NLI & Bilingual BERT & NL/SL & SYN, UniFreq & / \\
\hline
\multirow{2}{*}{\citet{lauscher_zero_2020}} & DP, POS, NER, NLI, QA & \multirow{2}{*}{mBERT, XLM-R} & \multirow{2}{*}{NL} & SYN, PHON, INV, GEN, GEO & \multirow{2}{*}{lang2vec} \\
\hline
\citet{dolicki2021} & \multirow{2}{*}{NER, NLI, POS} & \multirow{2}{*}{XLM-R} & \multirow{2}{*}{NL} & \multirow{2}{*}{GEN, GEO, SYN} & lang2vec, WALS \\
\hline
\multirow{2}{*}{\citet{srinivasan2021}} & \multirow{2}{*}{NER, NLI, POS} & \multirow{2}{*}{mBERT, XLM-R} & \multirow{2}{*}{NL} & \multirow{2}{*}{ALL} & lang2vec, WALS \\
\hline
\multirow{2}{*}{\citet{ahuja_multi_2022}} & DC,  NER, NLI, POS, QA & \multirow{2}{*}{mBERT, XLM-R} & \multirow{2}{*}{NL} & ALL, GEN, GEO, PHON, SYN & lang2vec, WALS \\
\hline
\multirow{2}{*}{\citet{deshpande_when_2022}} & NER, NLI, POS, QA & Bilingual RoBERTa (small) & \multirow{2}{*}{SL} & \multirow{2}{*}{SYN} & \multirow{2}{*}{/} \\
\hline
\citet{de_vries_make_2022} & POS & XLM-R Base & NL & FAM, SYN, WS, WST & / \\
\hline
\multirow{2}{*}{\citet{eronen_transfer_2022}} & \multirow{2}{*}{DC} & \multirow{2}{*}{mBERT, XLM-R} & \multirow{2}{*}{NL} & \multirow{2}{*}{ALL} & eLinguistics, WALS \\
\hline
\multirow{2}{*}{\citet{wu2022}} & AJ, SA, SS, NLI & \multirow{2}{*}{English RoBERTa} & \multirow{2}{*}{SL} & \multirow{2}{*}{SYN} & \multirow{2}{*}{/} \\

\hline
\end{tabular}


    \caption{List of studies investigating linguistic features that impact cross-lingual transfer. The \underline{\textbf{Lang. type}} column indicates the type of language that has been used. We use the following abbreviations. \textbf{NL}: Natural Languages, \textbf{SL}: Synthetic languages. The \underline{\textbf{Features}} column indicates which linguistic features have been investigated. We use the following abbreviations. \textbf{ALL}: Aggregated language distance of multiple linguistic features, \textbf{GEN}: Genetic distance, \textbf{GEO}: Geographical distance, \textbf{INV}: Inventory, \textbf{PHON}: Phonology, \textbf{SYN}: Syntax,  \textbf{UniFreq}: Unigram Frequency, \textbf{WS}: Writing system, \textbf{WST}: Writing system type. The \underline{\textbf{Metrics}} column indicates which type of metric has been used to measure language similarity between natural languages. The abbreviations of the \underline{\textbf{Task}} column can be found in Table \ref{tab: table_ALL} in Appendix \ref{sec:appendix}.
    }
    \label{tab:Linguistic_Similarity}
\end{table*}

The hypothesis that linguistic similarity correlates with cross-lingual transfer performance has been examined repeatedly. With regard to quantifying such a relationship, we observe two main approaches: \textbf{i)} synthetically modifying a specific linguistic feature of a natural language and observing the impact on transfer performance by controlling the magnitude of the modification; and \textbf{ii)} using linguistic similarity metrics to capture the similarity between two natural languages.

Two established linguistic similarity metrics which are commonly used for this purpose are: the World Atlas of Language Structures (WALS)\footnote{\url{https://wals.info/}} \citep{dryer_wals_2013}, a database of structural properties of languages, and \texttt{lang2vec}\footnote{\texttt{lang2vec} enables querying the URIEL database. It extracts vectors which encode different linguistic components for each language. This, in turn, allows to quantify the similarity or dissimilarity between languages.}, a tool providing vectors that represent linguistic properties of languages based on the URIEL \citep{littell_uriel_2017} database. An alternative metric for evaluating linguistic similarity is eLinguistics\footnote{\url{http://www.elinguistics.net/}} \citep{beaufils_stochastic_2020}, which is based on the comparison of consonants in word pairs. Table \ref{tab:Linguistic_Similarity} lists papers that have investigated the impact of linguistic similarity, along with the linguistic components that were studied and the metrics used.

\paragraph{Is Word Order Important?}
The impact of word order\footnote{\textit{Word order} describes the degree of similarity between the source and target language in terms of elements like subject-object-verb, subject-verb and object-verb order.}, or more generally, syntax, has been extensively investigated in the past. Based on experiments with different settings, its positive effect on cross-lingual transfer has been confirmed for Dependency Parsing (DP) \citep[e.g.,][]{lin_choosing_2019, lauscher_zero_2020}, Named Entity Recognition (NER) \citep[e.g.,][]{dolicki2021, deshpande_when_2022, ahuja_multi_2022}, Part-Of-Speech Tagging (POS) \citep[e.g.,][]{ahuja_multi_2022, de_vries_make_2022, deshpande_when_2022}, Natural Language Inference (NLI) \citep[e.g.,][]{k_cross-lingual_2020, lauscher_zero_2020, ahuja_multi_2022} and Question Answering (QA) \citep[e.g.,][]{deshpande_when_2022, ahuja_multi_2022, lauscher_zero_2020}. Furthermore, \citet{dufter_identifying_2020} sought to validate these findings on a representation level by evaluating cross-lingual transfer on word translation, word retrieval and sentence retrieval.

Despite the common findings stated above, there are contradictions in the results of a number of studies in which different experimental settings are used.
\citet{wu2022} and \citet{deshpande_when_2022} investigated the impact of word order by isolating it from other factors. In both works, language variants were created by randomly permutating, inversing, or consistently adapting word order to a different language via a dependency tree. A common finding has been that reversed or randomized word order deteriorates cross-lingual transfer performance significantly more than adapting the word order to a different language. This makes it hard to compare the aforementioned findings to results from \citet{dufter_identifying_2020} and \citet{k_cross-lingual_2020} who solely evaluated on language variants with reversed or randomly permuted word order, respectively. Even if both latter works found evidence that word order impacts transfer performance, it is important to consider that \citet{wu2022} and \citet{deshpande_when_2022} have comparable findings in similar settings but observed a less significant effect when switching to a more structured syntactic modification.

On the other hand, \citet{lauscher_zero_2020} and \citet{ahuja_multi_2022} obtained results containing evidence that word order may be more important for mBERT than for XLM-R. A possible explanation for this finding is that mBERT encodes more syntactic knowledge than XLM-R, as shown by \citet{zheng_probing_2022}. 

\paragraph{Which Other Linguistic Features Affect Cross-Lingual Transfer?}
In addition to examining the effect of similar word order, some research has also focused on the impact of other linguistic characteristics. \citet{srinivasan2021} measured general language similarity by aggregating multiple lang2vec vectors. They observed a high, medium and low importance of language similarity for cross-lingual transfer in POS, QA and NLI, respectively. Their observation holds for both mBERT and XLM-R. By evaluating on a document classification task, \citet{eronen_transfer_2022} observed a medium correlation between the cross-lingual transfer performance of both models and an aggregation of WALS features.

On a more detailed level, low \textbf{geographical distance}\footnote{\textbf{Geographical distance} is based on the orthodromic distance between languages' primary locations.} between languages has been found to be beneficial for cross-lingual transfer on several occasions \citep{lin_choosing_2019, lauscher_zero_2020, dolicki2021, ahuja_multi_2022}. Similarly, low \textbf{genetic distance}\footnote{\textbf{Genetic distance} between two languages measures their degree of common ancestry.} has also been shown to positively affect cross-lingual transfer \citep{lin_choosing_2019, lauscher_zero_2020, dolicki2021, de_vries_make_2022, eronen_transfer_2022}. However, it has not been selected as a predictive feature in the Lasso regression performed by \citet{ahuja_multi_2022}. Low \textbf{phonological distance}\footnote{\textbf{Phonological distance} measures the difference of phonological properties between languages.} has been demonstrated to be more important for token-level tasks (NER, POS, DP, QA) than for sentence-level tasks (NLI, MT) \citep{lin_choosing_2019, lauscher_zero_2020, ahuja_multi_2022}. \textbf{Inventory features}\footnote{\textbf{Inventory features} describe a language's phonetic, phonological, and morphological components.} have been shown to be of low importance when selecting a suitable transfer language \citep{lin_choosing_2019, lauscher_zero_2020}.

Furthermore, \citet{k_cross-lingual_2020} investigated the utility of the hypothesis that similar words have a similar frequency in their respective language (Zipf's law). The authors assessed cross-lingual transfer using a synthetic target language, which has a similar unigram frequency but no other explicit commonality. Although its utility in combination with additional factors has not been evaluated, unigram frequency has been found to be unable to ensure a successful transfer between languages as a standalone feature.

\paragraph{Conclusion}
In previous research, syntax has been suggested as potentially the most important linguistic contributor for better cross-lingual transfer. However, we hypothesize that its impact may be overestimated when assessed by randomly permutating or inversing word order, since such syntactic modifications are unlikely to occur in natural languages. Besides syntax, other linguistic features, such as geographical, genetic and phonological similarity, have been identified as potential linguistic contributors as well. In addition, we emphasize the importance of investigating the distinct interplay of different linguistic features.

\subsection{Lexical Overlap}
Since lexical overlap may intuitively create a potential connection between closely related languages and therefore possibly explain the varying transfer performance across language pairs, its impact has been investigated on many occasions. Lexical overlap merely specifies the amount of shared words or subwords between a language pair. Typically, it is calculated as the percentage of unique words or subwords common to the vocabularies of both the source and target languages. There are various approaches to quantify lexical overlap between languages. A common corpus-based method is to divide the number of shared words or subwords between two monolingual corpora by the total number of unique words or subwords in both corpora. Two further metrics that aim to quantify lexical overlap are ezGlot\footnote{\url{https://www.ezglot.com/}} \citep{kovacevic_ezglot_2022} and the normalized Levenshtein distance (LDND) \citep{wichmann_evaluating_2010}.

\paragraph{Does High Lexical Overlap Improve Cross-Lingual Transfer?}
While many studies have found a positive correlation between lexical overlap and cross-lingual transfer performance \citep{wu_beto_2019, patil_overlap-based_2022, de_vries_make_2022}, other studies do not support the existence of such a positive correlation \citep{pires_how_2019, tran_zero-shot_2019, k_cross-lingual_2020, conneau_emerging_2020}.

\citet{pires_how_2019}, \citet{tran_zero-shot_2019} and \citet{wu_beto_2019} applied the traditional cross-lingual zero-shot transfer evaluation pipeline (see Section \ref{cl_transfer_def}) on different tasks and natural languages. Besides showcasing the cross-lingual capacity of mBERT, their objective was to measure the impact of lexical overlap on this ability. Despite the similarities of their experiments, their findings are not all consistent. Based on the experiments on POS and DP in more than 16 languages, \citet{pires_how_2019} and \citet{tran_zero-shot_2019} have found that cross-lingual transfer performance is largely independent of lexical overlap. \citet{wu_beto_2019}, on the other hand, derived a correlation between transfer performance and lexical overlap from results on more tasks but fewer languages.

\citet{de_vries_make_2022} evaluated cross-lingual transfer performance across languages with different writing systems. They found that a shared writing system and thus a higher lexical overlap (measured by LDND) contribute to better cross-lingual transfer. However, they also showed that cross-script transfer is not impossible. Such a finding clearly supports the hypothesis that lexical overlap should not be seen as a self-contained factor. Based on these findings, it becomes evident that a more detailed analysis of the impact of lexical overlap is needed. Such detailed analyses would provide additional clarification on the apparent contradictions among past contributions.

\paragraph{Does the Impact of Lexical Overlap on Transfer Performance Depend on Other Linguistic Features?}
With the intention of a more fine-grained investigation, \citet{k_cross-lingual_2020} and \citet{conneau_emerging_2020} have conducted experiments in a controlled setup by synthetically adjusting the amount of lexical overlap. In both cases, no significant correlation between lexical overlap and transfer performance was observed. \citet{patil_overlap-based_2022} used similar configurations but differentiated between high- and low-resource settings. In contrast to previous findings, they observed a positive correlation between subword overlap and transfer performance. Furthermore, they concluded that this correlation increases when the source language has a smaller pre-training corpus.

\citet{deshpande_when_2022} took this a step further by transferring exclusively from synthetic English to English. This allowed them to isolate the impact of lexical overlap and control interactions with other linguistic features. From their experiments, it can be concluded that lexical overlap matters most when the word orders of the source and target languages differ. This finding explains the results of \citet{k_cross-lingual_2020} and \citet{conneau_emerging_2020} who only used language pairs of similar word order and did not observe a high impact of lexical overlap on transfer performance. The only language pair in their experiments with dissimilar word order was English-Hindi, which has small lexical overlap by default due to their different scripts. Consequently, further reducing the overlap is, as observed in their results, not expected to impact transfer performance. Moreover, this potentially explains the aforementioned findings of \citet{pires_how_2019} and \citet{tran_zero-shot_2019} who performed their experiments on a subset of languages for which word order and lexical overlap are strongly correlated. In both studies, language pairs with low lexical overlap were most likely also differing in their word order, while language pairs with higher lexical overlap tended to have similar word order. \citet{pires_how_2019}, unfortunately, did not provide exact transfer performance values. However, in line with our aforementioned observations, in their study a correlation between transfer performance and lexical overlap could be observed in language pairs with low lexical overlap and thus dissimilar word order. This correlation decreases as lexical overlap increases and thus word order becomes mostly similar.

\paragraph{Does the Impact of Lexical Overlap on Transfer Performance Depend on the Type of Downstream Task?}
\citet{lin_choosing_2019}, \citet{srinivasan2021} and \citet{ahuja_multi_2022} trained predictors to predict the cross-lingual transfer performance of a given language model for a variety of downstream tasks. Lexical overlap between source and target languages was selected as one of the predictor variables. By comparing the feature importance values of lexical overlap, clear differences across different types of downstream tasks emerged. While \citet{lin_choosing_2019} and \citet{srinivasan2021} observed high feature importance values of lexical overlap for syntactic tasks like POS, NER and DP, and lower feature importance values for the semantic-oriented task of NLI, \citet{ahuja_multi_2022} found the opposite. 

Given the minor but numerous differences among studies, providing a thorough explanation of the aforementioned contradictory findings is challenging. One notable distinction among the three similar contributions is the use of tree-based methods, specifically Gradient-Boosted Decision Trees and XGBoost, by \citet{lin_choosing_2019} and \citet{srinivasan2021}, respectively, and the use of Lasso Regression, a type of linear regression, by \citet{ahuja_multi_2022}. Given that tree-based models are able to capture nonlinear relationships between the dependent and independent variables while Lasso Regression can only describe such a relationship linearly, the latter method might attribute higher feature importance to linearly related predictors compared to predictors that have a more significant but nonlinear impact on the dependent variable. A recent study by \citet{patankar_train_2022} provides evidence in support of our hypothesis.

\paragraph{Conclusion}
We found evidence that lexical overlap is particularly important when the pre-training corpus for the source language is small or when the word order between the source and target languages is dissimilar. However, we conclude that lexical overlap is not a sufficient standalone factor to explain cross-lingual transfer. We also observed in experiment results in the literature that cross-lingual transfer is feasible between languages with different scripts (and thus zero lexical overlap), which further supports our conclusion. We recommend that future experiments take a closer look at the interaction between lexical overlap and further contributing factors. Moreover, future experiments may be set up in a way to provide additional insight into task-specific differences that are currently not fully understood.

\subsection{Model Architecture}
Model architecture may be crucial to the success of cross-lingual transfer because it determines how a model processes and represents information. Therefore, it is closely connected to the model's capacity to learn and capture knowledge. An ill-suited architecture could potentially hinder the model's ability to transfer knowledge from one language to another.

\paragraph{Which Model Architecture Components Can Affect Transfer Performance?}
\citet{k_cross-lingual_2020} provided one of the first investigations on the impact of model architecture on cross-lingual transfer. In their study, they focused on three main architectural components of Transformer-based models: \textbf{i)} network depth, \textbf{ii)} number of attention heads, \textbf{iii)} number of model parameters. They found that an increased network depth (i.e., more hidden layers), with a fixed number of model parameters, leads to better cross-lingual transfer. Increasing the number of model parameters with a fixed number of hidden layers had a similar but less significant impact. The number of attention heads, on the other hand, were found to be irrelevant for cross-lingual transfer performance. In their experiments, satisfactory transfer performance could even be achieved with only a single attention head.

\citet{conneau_emerging_2020} trained a bilingual BERT model where all parameters are shared, and compared the transfer performance to the case where the embedding layer and/or up to the first six Transformer layers are separated for both languages. In the experiments on NLI, DP, and NER for three different natural language pairs, they observed that the transfer performance decreases when fewer layers are shared. This finding led the authors to hypothesize that a limited model capacity requires the model to use its parameters more efficiently by aligning the representations of semantically similar text across different languages, instead of creating separate embedding spaces for different languages. This hypothesis was confirmed by \citet{dufter_identifying_2020} who observed a degradation of mBERTs cross-lingual transfer ability by purposely overparameterizing the model. On the other hand, the authors referred to the "curse of multilinguality" \citep{conneau_unsupervised_2020} which states that, for a fixed model size, the number of languages a model can cover until its overall performance starts to decrease is limited. This can be alleviated by expanding the model capacity, i.e., by increasing the number of parameters, but as mentioned previously, too many parameters could deteriorate cross-lingual transfer performance.

\citet{wu2022} demonstrated the importance of a well-trained embedding layer for cross-lingual transfer. When the embedding layer is reinitialized before fine-tuning, the performance on the GLUE benchmark \citep{wang_glue_2018} decreases by 40\%. More specifically, \citet{deshpande_when_2022} found that the cross-lingual alignment of the static token embeddings used by the embedding layer is crucial for satisfactory cross-lingual transfer performance.

\paragraph{Conclusion}
There is evidence to suggest that an overparameterized model might create language-specific sub-spaces and therefore struggle to provide cross-lingual representations. Concurrently, models with fewer parameters are required to use their parameters more efficiently and thus align representations across languages more easily. Therefore, we strongly suggest to explore how the trade-off between languages and parameters affects cross-linguality in MLLMs. 

Furthermore, one contribution has revealed evidence that for a fixed number of parameters, model depth can be more important than the number of attention heads. However, it is not well studied yet how model architecture components and data-specific components (e.g., dataset size, number of languages) interact to impact cross-lingual transfer performance.

\subsection{Pre-Training Settings}
Given that MLLMs are able to perform zero-shot cross-lingual transfer, their cross-lingual capacity has to emerge during pre-training as they are not exposed to any task-specific data in the target language during fine-tuning. Therefore, investigating factors related to the pre-training process could lead to a better understanding of the cross-lingual capacity of MLLMs as well as how to further improve it.

\paragraph{Which Pre-Training Components Contribute to the Cross-Lingual Capabilities of MLLMs?}
\citet{devlin_bert_2019} introduced the Next Sentence Prediction (NSP) objective to pre-train language models in combination with the Masked Language Model (MLM) objective. However, the usefulness of NSP for downstream tasks has been debated on several occasions \citep{yang_xlnet_2019, conneau_cross-lingual_2019, liu_roberta_2019, joshi_spanbert_2020}. \citet{k_cross-lingual_2020} probed its impact on cross-lingual transfer performance. By removing NSP from the pre-training process, performance improved for both NER and NLI. This finding is particularly remarkable for NLI as this task is considered to be closely related to NSP, as both tasks involve the classification of sentence pairs.
Furthermore, they also found that training on subwords rather than words or characters provides more cross-lingual capacity to the model. Lastly, it has been shown that adding a language identity marker to the input during pre-training does not significantly improve cross-lingual transfer performance. This outcome may suggest that MLLMs automatically learn language-specific information \citep{wu_beto_2019, liu2020} or that such additional input is not necessary for their cross-lingual capability. Furthermore, \citet{liu2020} showed that pre-training on longer input sequences helps MLLMs to achieve better cross-lingual transfer abilities, especially when pre-trained on large corpora.

Apart from the learning objective, the impact of tokenizers and their vocabulary on a model's cross-lingual potential have been examined as well. \citet{artetxe_cross-lingual_2020} evaluated transfer performance of bilingual and multilingual BERT models pre-trained with different vocabulary settings on four different downstream task datasets. In multilingual settings, they found that increased joint vocabulary size\footnote{Experiments were conducted with vocabulary sizes of 32k, 64k, 100k, and 200k.} leads to improved cross-lingual transfer performance. Furthermore, in the context of bilingual models, cross-lingual transfer performance is enhanced when disjoint subword vocabularies\footnote{A joint vocabulary of 32k subwords was compared to two separate vocabularies, each with 32k subwords, for each language.} are utilized instead of a joint subword vocabulary for both languages. That said, it is unclear how well disjoint vocabularies would perform when scaling the model to more languages. 

\citet{ahuja_multi_2022} also studied the effect of tokenizers on cross-lingual transfer. They quantify tokenizer quality by applying two metrics introduced by \citet{rust_how_2021}, namely the tokenizer's \textit{fertility} and its proportion of continued words. Both features are included in their cross-lingual transfer performance prediction model. By looking at the feature importance values, it became clear that cross-lingual transfer performance depends significantly more on a high-quality tokenizer for POS, NER and QA than for Document Classification (DC) and Sentence Retrieval (SR). Such a finding aligns with the fact that the former downstream tasks operate to a greater extent on token level than the latter ones.

\paragraph{Conclusion}
Previous studies have identified a number of pre-training components which may enable an improved cross-lingual transfer capacity of MLLMs. Some examples include removing NSP from the pre-training learning objective, a larger vocabulary size and a high-quality multilingual tokenizer.

\subsection{Pre-Training Data}
MLLMs, such as mBERT, are able to learn cross-lingual representations during pre-training without having been specifically designed to do so. This may happen as a result of the model's exposure to multiple languages during the pre-training phase. However, the impact of the pre-training corpus on this self-learned ability is not yet fully comprehended.

\paragraph{Does the Pre-training Corpus Size Influence a Model’s Cross-Lingual Transfer Ability?}
\citet{lauscher_zero_2020}, \citet{srinivasan2021} and \citet{ahuja_multi_2022} found that the size of the pre-training target language corpora correlates strongly with the transfer performance of mBERT and XLM-R for high-level tasks (NLI \& QA) and less for low-level tasks (DP, POS, NER).

\citet{liu2020} performed a more controlled experiment by comparing two multilingual BERT models pre-trained on different amounts of data from 15 languages. When trained on a small corpus of 200k sentences per language, mBERT showed poor zero-shot cross-lingual transfer performance, with results only comparable to those of non-contextualized word embedding models such as GloVe \citep{pennington_glove_2014} and Word2Vec \citep{mikolov_efficient_2013} that were also trained on the same amount of data. Increasing the pre-training corpus size to 1000k sentences per language resulted in significantly improved transfer performance of mBERT, while both non-contextualized word embedding models did not demonstrate such an enhancement in transfer performance.

\citet{lin_choosing_2019} found that the ratio between the pre-training data corpus size of the transfer and target language is an important factor for successful cross-lingual transfer for POS but less so for MT and DP. However, the size of the target language pre-training corpus is not examined as a distinct feature in their work, making it more challenging to compare their findings with those mentioned previously.

\paragraph{Does the Source of the Pre-training Corpus Affect Cross-Lingual Transfer Performance?}
\citet{dufter_identifying_2020} found that cross-lingual transfer performance decreases when the respective monolingual pre-training corpora come from the same domain but are not parallel (e.g., by pre-training on different parts of the same corpus from a given domain). \citet{conneau_emerging_2020} obtained similar results for monolingual pre-training corpora from different domains (e.g., Wikipedia vs. Common Crawl). \citet{deshpande_when_2022} found that pre-training on corpora from different domains has a more significant negative impact on cross-lingual transfer performance than pre-training on non-parallel corpora from the same domain. Interestingly, \citet{conneau_emerging_2020} and \citet{deshpande_when_2022} found that the negative effect of different pre-training corpora sources on cross-lingual transfer performance is the most significant for NER. A potential explanation could be that in both cases, the NER dataset consists of Wikipedia text which was also used as the pre-training corpus in their baseline experiments. To the best of our knowledge, there is no research available on the impact of using a shared source for pre-training and task-specific data in the cross-lingual transfer context.

\paragraph{Conclusion}
Target language pre-training corpus size and comparable corpora sources across languages  have been identified as two crucial factors for enhanced cross-lingual transfer capabilities in MLLMs. However, pre-training corpus size of the target language has been shown to be more important for higher-level than for lower-level tasks.

\section{Related Work}
Recently, numerous studies have investigated how to leverage the cross-lingual potential of MLLMs for better transfer among languages. \citet{pikuliak_cross-lingual_2021} conducted a survey on existing cross-lingual transfer paradigms but did not investigate the components that are responsible for their inner workings. \citet{doddapaneni2021}, in their survey on pre-trained MLLMs, commented on various factors that affect cross-lingual transfer. Since they discussed a wide range of topics, they could not investigate in depth the findings from the studies that examined these factors. After the publication of that work, many studies have further investigated various factors that impact transfer performance and have helped to resolve some of the conflicts among past contributions.

\citet{malkin_balanced_2022} introduced a \textit{Linguistic Blood Bank} that shows that not all languages transfer equally well among each other. This emphasizes the need for a clearer understanding of the underlying factors that contribute to this imbalance. On a related note, \citet{turc2021} found that English is not the overall best source language for cross-lingual transfer, despite its dominance in the pre-training corpus.

Hence, automating the process of selecting a source language for cross-lingual transfer has been pursued on many occasions \citep{lin_choosing_2019,lauscher_zero_2020, srinivasan2021, dolicki2021}. These attempts focused on creating meta-models\footnote{In this context, the objective of a meta-model is to predict the performance of other models.} which aim to predict the most suitable source language for a given use-case based on some of the factors from Section \ref{factors}.

By incorporating typological features, \citet{ansell_mad-g_2021}, \citet{lee_fad-x_2022} and \citet{chronopoulou_language-family_2023} enhanced the performance of adapters for low-resource languages. However, our survey reveals that adapters and other methods could benefit from more than just typological factors when dealing with low-resource scenarios.

\section{Discussion}
Building on previous research, our study investigated various factors that impact cross-lingual transfer performance. We examined a range of factors, including language-related factors as well as factors related to the models and training data. One of the existing challenges is the presence of contradictory findings from past studies. To better understand these discrepancies, we outlined possible explanations that could account for these differences, including the varying implementation details of experiments and evaluation methods.

One of the key variations among the various studies is the use of synthetic and natural languages. Synthetic languages can be created with a controlled level of variation by manipulating specific linguistic features. However, they may not capture the full range of complexity found in natural languages, which may limit their usefulness in drawing conclusions that apply to real-world settings.

While we acknowledge the value of the efficiency of using transfer performance prediction models to automate the selection of transfer languages, the accuracy of relying on feature importance values to make conclusions about the individual impact of specific factors on cross-lingual transfer performance cannot be taken as an absolute.

Our survey results show that all the factors we examined affect cross-lingual transfer in different ways and settings. Although the interaction of factors has only been investigated in a limited number of past studies, our findings suggest that some factors can influence the importance of others. Additionally, there is evidence suggesting that there are task-specific differences, for example, the pre-training corpus size being more important for higher-level tasks and lexical overlap, and word order being more important for lower-level tasks. Therefore, we strongly encourage future research to examine the full range of interactions among different factors as well as the underlying reasons for task-specific divergences.

Given that especially linguistic features have been shown to have a strong impact on cross-lingual transfer performance, we suggest that future research could examine whether languages are indeed the most suitable basis for constructing multilingual models. Instead of focusing on the distribution of languages in the pre-training corpus, it might be more efficient to focus on the distribution of linguistic features. One possible approach is to cluster texts according to their syntactic complexity or their morphological diversity, irrespective of their language affiliation. This would enable the development of a model that could potentially better transfer to languages that were absent in the pre-training corpus but which share linguistic features with the languages that the model has seen during pre-training.

In addition, we advocate for the development of more multilingual downstream task datasets that encompass a wider and more diverse range of languages, as this would enable a more comprehensive and robust assessment of cross-lingual transfer capabilities across various language models and approaches. Furthermore, we urge more investigation on the influence of the aforementioned factors on generative models, as this area remains relatively unexplored despite the current prominence of GPT-like models.

\section*{Limitations}
One potential limitation of this review is our selection bias which may affect the representativeness of the included papers. Another limitation is the potential differences in methodologies across the papers we reviewed, which makes it difficult to draw generalizable conclusions. Different studies use different experimental settings and methods for measuring feature importance, which could also impact the comparability of the findings across the included studies. Furthermore, we acknowledge the potential publication bias which might lead to an overestimation of the impact of different factors, as studies with statistically significant results may be more likely to be published than those with non-significant results.

\section*{Ethics Statement}
We have carefully reviewed the relevant literature to ensure that all research included in this review has been conducted in accordance with ethical guidelines. We have also attempted to present a fair and accurate representation of the current state of research on this topic. We hope that this review will contribute to the ongoing debate about the factors impacting cross-lingual transfer performance, with the ultimate goal of ensuring that low-resource languages can equally benefit from the use of multilingual language models. We believe that it is important for all languages and communities to have equal access to the benefits and opportunities provided by the advances in natural language processing, and we hope that our review will serve as a useful resource in this regard.

\bibliography{zotero}
\bibliographystyle{acl_natbib}

\appendix

\section{Appendix}
\label{sec:appendix}

\begin{table*}
\centering
\rotatebox{90}{
\begin{tabular}{lllp{0.8cm}l}
\hline
\multirow{2}{*}{\textbf{Paper}} & \multirow{2}{*}{\textbf{Task}} & \multirow{2}{*}{\textbf{Model}} & \textbf{Lang. type} & \multirow{2}{*}{\textbf{Factor}}\\
\hline
\citet{lin_choosing_2019} & DP, MT, POS, EL & / & NL & LO, LS, PTD \\
\citet{pires_how_2019} & POS, NER & mBERT & NL & LO, LS \\
\citet{tran_zero-shot_2019} & DP & mBERT & NL & LO, LS \\
\citet{wu_beto_2019} & DC, NER, DP, NLI, POS & mBERT & NL & LO \\
\citet{artetxe_cross-lingual_2020} & NLI, DC, QA &  Bilingual BERT, mBERT & NL & PTS \\
\citet{conneau_emerging_2020} & NLI, NER, DP & Bilingual BERT & NL/SL & LO, MA, PTD \\
\citet{dufter_identifying_2020} & WA, WT, SR & BERT (small) & SL & LS, MA, PTD \\
\citet{lauscher_zero_2020} & DP, POS, NER, NLI, QA & mBERT, XLM-R & NL & LS, PTD \\
\citet{liu2020} & NLI & mBERT & NL & PTD, PTS \\
\citet{k_cross-lingual_2020} & NLI, NER & Bilingual BERT & NL/SL & LO, LS, MA, PTS \\
\citet{dolicki2021} & NLI, NER, POS & XLM-R & NL & LS \\
\citet{srinivasan2021} & NLI, NER, POS & mBERT, XLM-R & NL & LO, LS, PTD \\
\citet{wu2022} & SA, AJ, SS, NLI & English RoBERTa & SL & LS, MA \\
\citet{ahuja_multi_2022} & DC, NLI, POS, NER, QA & mBERT, XLM-R & NL & LO, LS, PTD, PTS \\
\citet{de_vries_make_2022} & POS & XLM-R Base & NL & LO, LS \\
\citet{deshpande_when_2022} & NLI, NER, POS, QA & Bilingual RoBERTa (small) & SL & LO, LS, MA, PTD \\
\citet{eronen_transfer_2022} & DC & mBERT, XLM-R & NL & LS \\
\citet{patil_overlap-based_2022} & NER, POS, DC, NLI & mBERT (12 languages) & NL/SL & LO \\
\hline
\end{tabular}

}
\caption{\label{tab: table_ALL}
List of studies investigating factors that impact cross-lingual transfer. The \underline{\textbf{Task}} column indicates the downstream tasks that experiments have been performed on. We use the following abbreviation: \textbf{AJ}: Acceptability Judgement, \textbf{DC}: Document Classification, \textbf{DP}: Dependency Parsing, \textbf{EL}: Entity Linking, \textbf{LID}: Language Identification, \textbf{LS}: Language Similarity, \textbf{MTQE}: Machine Translation Quality Estimation, \textbf{NER}: Named Entity Recognition, \textbf{NLI}: Natural Language Inference, \textbf{POS}: Part-of-speech tagging, \textbf{QA}: Question Answering, \textbf{SA}: Sentiment Analysis, \textbf{SR}: Sentence Retrieval, \textbf{SS}: Sentence similarity, \textbf{WA}: Word Alignment, \textbf{WT}: Word Translation. The \underline{\textbf{Model}} column indicates the models that were employed in the experiments of each paper. The \underline{\textbf{Factor}} column indicates the factors for cross-lingual transfer ability that have been investigated. We use the following abbreviations: \textbf{LO}: Lexical Overlap, \textbf{LS}: Language Similarity, \textbf{MA}: Model Architecture, \textbf{PTS}: Pre-Training Settings, \textbf{PTD} : Pre-Training Data. The \underline{\textbf{Lang. type}} column indicates the type of language that has been used. We use the following abbreviations. \textbf{NL}: Natural Languages, \textbf{SL}: Synthetic languages}
\end{table*}

\end{document}